\documentclass[sigconf,nonacm]{acmart}

\usepackage{balance}
\renewcommand\footnotetextcopyrightpermission[1]{} 
\AtBeginDocument{%
  }

\begin{document}

\title[Multimodal Sentiment Analysis]{Hybrid Multimodal Feature Extraction, Mining and Fusion for Sentiment Analysis}



\author{Jia Li}
\authornote{Team leader}
\affiliation{%
  \institution{School of Computer Science and Information Engineering, \\ Hefei University of Technology}
  \city{Hefei}
  \country{China}
}
\email{jiali@hfut.edu.cn}

\author{Ziyang Zhang}
\affiliation{%
  \institution{School of Computer Science and Information Engineering, \\ Hefei University of Technology}
  \city{Hefei}
  \country{China}
}
\email{2021171178@mail.hfut.edu.cn}

\author{Junjie Lang }
\affiliation{%
  \institution{School of Computer Science and Information Engineering, \\ Hefei University of Technology}
  \city{Hefei}
  \country{China}
}
\email{2020171152@mail.hfut.edu.cn}

\author{Yueqi Jiang}
\affiliation{%
  \institution{School of Computer Science and Information Engineering, \\ Hefei University of Technology}
  \city{Hefei}
  \country{China}
}
\email{2017217795@mail.hfut.edu.cn}

\author{Liuwei An}
\affiliation{
  \institution{School of Computer Science and Information Engineering, \\ Hefei University of Technology}
  \city{Hefei}
  \country{China}
}
\email{superalw@163.com}

\author{Peng Zou}
\affiliation{
  \institution{School of Computer Science and Information Engineering, \\ Hefei University of Technology}
  \city{Hefei}
  \country{China}
}
\email{zonepg666@gmail.com}

\author{Yangyang Xu}
\affiliation{
  \institution{School of Computer Science and Information Engineering, \\ Hefei University of Technology}
  \city{Hefei}
  \country{China}
}
\email{1798265271@qq.com}

\author{Sheng Gao}
\affiliation{
  \institution{School of Computer Science and Information Engineering, \\ Hefei University of Technology}
  \city{Hefei}
  \country{China}
}
\email{gaos@mail.hfut.edu.cn}

\author{Jie Lin}
\affiliation{
  \institution{School of Computer Science and Information Engineering, \\ Hefei University of Technology}
  \city{Hefei}
  \country{China}
}
\email{848284149@qq.com}

\author{Chunxiao Fan}
\affiliation{
  \institution{School of Computer Science and Information Engineering, \\ Hefei University of Technology}
  \city{Hefei}
  \country{China}
}
\email{fanchunxiao@hfut.edu.cn}

\author{Xiao Sun}
\authornote{Corresponding author}
\affiliation{%
  \institution{School of Computer Science and Information Engineering, \\ Hefei University of Technology}
  \institution{ZhongJuYuan Intelligent Technology Co., Ltd}
  \city{Hefei}
  \country{China}
}
\email{sunx@hfut.edu.cn}

\author{Meng Wang}
\authornotemark[2]
\affiliation{%
  \institution{School of Computer Science and Information Engineering, \\ Hefei University of Technology}
  \institution{Institute of Artificial Intelligence, Hefei Comprehensive National Science Center}
  \city{Hefei}
  \country{China}
}
\email{eric.mengwang@gmail.com}

\renewcommand{\shortauthors}{Jia Li et al.}

\begin{abstract}
  In this paper, we present our solutions for the Multimodal Sentiment Analysis Challenge (MuSe) 2022, which includes MuSe-Humor, MuSe-Reaction and MuSe-Stress Sub-challenges. The MuSe 2022 focuses on humor detection, emotional reactions and multimodal emotional stress utilizing different modalities and data sets. In our work, different kinds of multimodal features are extracted, including acoustic, visual, text and biological features. These features are fused by TEMMA and GRU with self-attention mechanism  frameworks. In this paper, 1) several new audio features,  facial expression features and paragraph-level text embeddings are extracted for accuracy improvement. 2) we substantially improve the accuracy and reliability of multimodal sentiment prediction by mining and blending the multimodal features. 3) effective data augmentation strategies are applied in model training to alleviate the problem of sample imbalance and prevent the model from learning biased subject characters. For the MuSe-Humor sub-challenge, our model obtains the AUC score of 0.8932. For the MuSe-Reaction sub-challenge, the Pearson's Correlations Coefficient of our approach on the test set is 0.3879, which outperforms all other participants. For the MuSe-Stress sub-challenge, our approach outperforms the baseline in both arousal and valence on the test dataset, reaching a final combined result of 0.5151.

\end{abstract}

\maketitle

\section{Introduction}

Emotion is an important information that people transmit in the process of communication. The change of emotional state affects people's perception and decision-making. Sentiment
analysis is an important research field of pattern recognition, which introduces emotional dimension into human-computer interaction. The modes of emotional expression include facial expression, speech, posture, physiological signals, words, etc. emotion recognition is essentially a problem of multimodal fusion. In this paper, we extensively present our solutions for the Multimodal Sentiment Analysis Challenge (MuSe) 2022 \cite{Christ22-TM2,Amiriparian22-TM2}, which includes three sub-challenges of Humor, Emotional Reactions, and Stress.

In the Humor Detection Sub-Challenge (MuSe-Humor), it aims to detect the presence of humor in football press conference recordings. Humor detection has been a hot topic in the field of natural language processing. The analysis of the level of humor in text can help to better implement tasks such as human-computer interaction. However, research on multimodal humor detection is relatively rare, in large part because of the difficulty of collecting and labeling multimodal humor datasets. In real scenarios, humor is not only reflected in a text modality, but also visual and acoustic modalities. For example, when a person is joking, he/she tends to smile and emits laughter. If the information from multiple modalities can be effectively used for humor detection, it can certainly improve the detection effect significantly.

In the Emotional Reactions Sub-Challenge (MuSe-Reaction), we need to predict the intensities of seven self-reported emotions from user-generated reactions to emotionally evocative videos. Everyone has their subjective feelings and expresses their emotions through behaviors such as facial expressions, words and body movements. Multimodal emotion recognition is to identify and predict emotions through these physiological responses and behavioral responses (ie, multimodal information). Due to the lack of robustness and low recognition rate of single-modal emotion recognition, the emotion recognition performance will be significantly reduced when the emotion signal is masked subjectively by humans or the emotion signal of a single channel is affected by other signals. For example, facial expressions are easy to be blocked, voice are easily disturbed by noise. 
Multimodal emotion recognition recognizes human emotional state by analyzing speech signals, visual signals and physiological signals, and uses the complementarity between multi-channel emotional information to improve the accuracy of emotion recognition.

In the Emotional Stress Sub-challenge (MuSe-Stress), the level of valence and psycho-physiological arousal in a time-continuous manner needs to be predicted from audio-visual recordings. It is necessary to predict the values of valence and arousal in a continuous way. Both valence and arousal have the problem of over fitting, especially the serious over fitting of arousal. There is a huge difference between the results of the development  set and the test set, so it is difficult to deal with this problem.


The main contribution of the proposed method can be summarized as:

1. We have tried to extract new audio features, new facial expression features and paragraph-level text embeddings,  directly leading to significant accuracy improvements to both the baseline model and our inference model;

2. We explore different modality effects for all the three sub-challenges of MuSe 2022 and found that some modalities play a negative role in one task while playing a positive role in other tasks. We substantially improve the accuracy and reliability of the  baseline approach for multimodal sentiment prediction by effectively making different modal features work collaboratively and mine them using self-attentive deep networks;

3. During model training, we employ effective data augmentation strategies to alleviate the problems of sample imbalance and prevent the model from learning biased subject characters, and successfully improve the generalization accuracies on the unseen test data.

The remainder of this paper is organized as follows. Related works are introduced in Section \ref{relate}. Section \ref{feature} describes feature extraction for the sentiment analysis. In Section \ref{fusion}, we present the details of the feature fusion framework. Section \ref{exp} presents the implementation and experiments to evaluate the proposed method. Our work is concluded in Section \ref{con}.\par

\section{Related Work}
\label{relate}

\subsection{Multimodal Features}

Multimodal features, such as visual features, audio features, text features and physiological signals have been well introduced in MuSe and AVECs. Participants can get a better performance in these emotional computing tasks by capturing the details of these multimodal features. 

In the visual modality, the facial expression is an important aspect to understand and analyze emotions. The Facial Action Coding System (FACS) proposed by Ekman and Friesen \cite{ekman1978facial} has been widely used in some studies. This method recognizes specific emotions based on facial Action Units (AU). In addition, geometric features based on multi-scale and multi-directional Gabor wavelet representation \cite{zhang1999feature} are also used for emotion recognition. With the wide application of deep learning, people find that the features based on deep learning can achieve better results. Poria et al. \cite{poria2016convolutional} propose a convolutional recurrent neural network to extract visual features in 2016, which uses CNN and RNN stacking training. In the past series of AVECs, participants use the deep learning method \cite{chen2017multimodal, huang2017continuous}, which is better than the traditional handcrafted features. Shizhe Chen et al. \cite{vaswani2017attention} propose that using convolutional neural networks for feature extraction can achieve better results. Sijie Mai \cite{mollahosseini2017affectnet} et al. propose a multimodal feature fusion strategy named partition-conquer-combine. In the recent AVEC 2019, Baltru{\v{s}}aitis et al. propose the use of 2D+1D convolutional neural networks \cite{baltruvsaitis2016openface},  which also proves that the learned audiovisual features can be used to improve performance.

In audio modality, prosodic features are widely used in emotion recognition tasks. Some acoustic features, such as Mel Frequency Cepstral Coefficients (MFCC), Spectral centroid, the Perceptual Linear Predictive Coefficients (PLP), have been widely used in these tasks and achieve great performance. These acoustic features can be extracted through opensmile \cite{eyben2009openear}, which is a popular audio feature extraction toolkit. Similar to visual features, deep learning has also been widely used in acoustic feature extraction. In AVEC 2018 and 2019, different methods \cite{zhao2018multi, chen2019efficient} are proposed by participants and prove the effectiveness of deep learning in acoustic feature representation learning. Text modality can also be used to solve problems such as emotion recognition. Word2Vec \cite{mikolov2013distributed} and GloVe \cite{pennington2014glove} are proposed, which further improves the effect of text modality. Recently, BERT and other models have been proposed and achieved excellent results in Natural Language Processing tasks, which are pre-trained with large amounts of text data.

\subsection{Model Structure}

As a traditional machine learning algorithm, Support Vector Machine (SVM) is applied to emotion computing tasks. However, it does not take the temporal information into account.
In recent years, the recurrent neural network (RNN) performs well in building temporal relationships, which has a great impact on emotion recognition and sentimental analysis tasks. The GRU and LSTM are widely used for continuous emotion recognition,     \cite{satar2022rome} proposes a novel mixture-of-expert transformer RoME that disentangles the text and the video into three levels: the roles of spatial contexts, temporal contexts, and object contexts. \cite{lin2022STVGFormer} proposes a concise and effective framework named STVGFormer, which models spatiotemporal visual-linguistic dependencies with a static branch and a dynamic branch. Google proposes the Transformer  \cite{chen2017multimodal} model and it can achieve remarkable results. Transformer utilizes self-attention and multi-head attention to model the temporal dependencies of different positions in a sequence, regardless of their distances.

\subsection{Multimodal Fusion}

Multi-modal fusion is an important strategy to solve multimodal tasks. Common fusion strategies include early fusion, late fusion, hybrid fusion and so on.

In the early fusion, video features, audio features, and text features are directly combined as general feature vectors for model analysis. Chen et al. \cite{chen2017multimodal} uses the early fusion strategy to fuse the features of different modalities and got a good result in AVEC 2017. \cite{Sebastian2019Fusion} utilizes early fusion and fed the extracted features through a convolutional neural network.

In the late fusion, the features of each modality are analyzed independently and the final results are fused. Glodek et al. \cite{glodek2013kalman} uses Kalman filter, which is a famous decision fusion method, to fuse the results of basic classifiers.  \cite{Paw2022Does} discusses the late fusion, the early fusion, and the sketch, and explains the importance of choosing the appropriate modal late fusion.

Hybrid fusion is a combination of early fusion and late fusion. Wollmer et al. \cite{wollmer2013youtube} proposes a hybrid fusion method. In \cite{wollmer2013youtube}, audio and visual features are fused at the feature level. And then decision level fusion is used to fuse the fusion result with the prediction of text classifier.

\section{Feature Extraction}
\label{feature}

\subsection{Acoustic Features}

\textbf{eGeMAPS}: The extended Geneva Minimalistic Acoustic Parameter Set (eGeMAPS) \cite{zhao2018multi} is an extension of GeMAPS. The 62 features of the original GeMAPS plus 26 extended features yield 88 acoustic features specifically designed for the speech emotion recognition task, which are the feature set of eGeMAPS. In this method, we use a different window size and hop size for re-extraction by the openSMILE toolkit \cite{chen2019efficient}. In the MuSe-Humor Challenge, the Passau-SFCH dataset contains data from only 10 coaches, and the tonal personality characteristics of each coach are probably causing the bias of the model learning direction. So we intend to blur out the personality characteristics in the original audio as much as possible, and adopt the data enhancement tool provided in \cite{papakipos2022augly}. We perform a pitch shift operation on the original audio.

\textbf{DeepSpectrum}: DeepSpectrum \cite{amiriparian2017snore} is a deep feature extraction of spectrograms (e.g., Mel-spectrograms) obtained from audio signals using pre-trained image Convolutional Neural Networks (CNNs). DenseNet121 \cite{huang2017densely} pre-trained on ImageNet \cite{russakovsky2015imagenet} is used as the CNN backbone to obtain 1024-dimensional feature vectors. We adopt the feature provided by the organizers of MuSe 2022 and preprocessed them to align them with other features in the temporal dimension.

\textbf{IS09}: The INTERSPEECH 2009 (IS09) feature set is presented at the INTERSPEECH 2009 Emotional Challenge \cite{schuller2009interspeech}. It contains 384 features as statistical functionals applied to low-level descriptor contours. We use the openSMILE toolkit \cite{schuller2013interspeech} to extract it.

\textbf{IS13}: To reflect more potential information in the audio signal, we extract the INTERSPEECH 2013 (IS13) feature presented on the INTERSPEECH 2013 Computational Paralinguistics Challenge \cite{kong2020panns} using the openSMILE toolkit. It contains 6373-dimensional feature vectors.

\textbf{MFCCs}: Mel-frequency cepstral coefficients (MFCCs) features are common audio features in speech recognition and emotion recognition, and it is very close to the human auditory system. We use the Python toolkit Librosa to extract 40-dimensional features, and calculate their deltas and delta-deltas, finally combine them to get a 120-dimensional feature vector.

\textbf{CNN14}: To obtain the high-level deep acoustic representations, we use a supervised model PANNs \cite{gemmeke2017audio} pre-trained on the AudioSet dataset \cite{2017Audio}. PANNs contains many different systems. We use the CNN14 system trained with 16 kHz audio recordings and get a 2048-dimensional feature vector.

\subsection{Visual Features}
VGGFace2 is a face recognition dataset, and models trained on it can encode the general facial features which have not a close relationship with expressions. FAUs is a traditional and effective approach to recording expressive features which is closely related to the field of facial expression recognition and affect computing. We take VGGFACE2 and FAUs provided by the organizers.


The ResNet deep neural network is proposed to solve the gradient disappearance or gradient explosion problem of the deep model. The ResNet network uses 4 modules composed of residual blocks, and each module uses several residual blocks with the same number of output channels. We  train \textbf{ResNet-18 using the mini-AffectNet dataset} selected from AffectNet \cite{mai2019divide}. Compared to the original AffectNet dataset, mini-AffectNet contains fewer but higher quality images and includes both regular human faces and precise facial expressions. To free the model from irrelevant background information and focus more on facial information, we use the OpenFace \cite{chen2019efficient} open source framework to detect and align faces and ResNet-18 to extract facial expression features. We normalize the face before training, including removing the outer parts of the face and rotating the face to keep the line between the eyes horizontal. Finally, targeting seven facial expression classifications, the pre-trained ResNet-18 model achieves an average accuracy of 65.27\%.

\subsection{Text Features}
\textbf{BERT feature}: We use the BERT features provided by the MuSe competition organizers, which are sentence-level features and word-level features, respectively. BERT \cite{devlin2018bert} is a pre-trained model proposed by Google AI. BERT set SOTA performance in 11 different NLP task. The German version of the pre-trained model was used in the MuSe Humor Challenge. The final obtained embedding vector is 768 dimensions.

\textbf{Sentence-BERT feature}: Sentence BERT is a framework for embedding sentences. When using BERT to embed sentences, it is common to either average the output layer results or take the output of the first token (the [CLS] token) as the sentence embedding result. The experimental results in \cite{reimers2019sentence} show that such approaches are usually unsatisfactory. Thus, Reimers et al. proposed Sentence-BERT \cite{reimers2019sentence}, Sentence BERT achieves outstanding results on semantic similarity tasks. Due to the specificity of German, we chose the Multi-Lingual pre-training model.

\textbf{Phrase feature}: The embeddings of both BERT and Sentence BERT are single-sentence level embeddings. According to Semantic Script Theory of Humor (SSTH) proposed by Raskin \cite{raskin1979semantic}, there is a contextual transitive relationship between punchline and context of the joke.  Annamoradnejad et al. \cite{annamoradnejad2020colbert} build a humor dataset containing 200K English short texts and propose an effective humor text detection model. The Multi-Lingual Model of Sentence BERT can map texts in different languages with the same semantic meaning into a similar semantic space, so we use Sentence BERT to embed the English humor dataset in \cite{annamoradnejad2020colbert} and retrain a humor text detection model using the model structure proposed in \cite{annamoradnejad2020colbert}. In addition, we use this pre-trained model to process the Passau-SFCH dataset, and finally generate a 256-dimensional feature vector.

\subsection{Biological Features}
\textbf{ECG, RESP and BPM Feature}: Three biological signals (i.e., Electrocardiogram (ECG), Respiration (RESP), and heart rate (BPM)) are captured at a sampling rate of 1 kHz.

\section{Methods}
\label{fusion}

\subsection{MuSe-Humor and MuSe-Reaction Sub-Challenges}

\begin{figure*}[htb]
	\centering
	\includegraphics[width=0.9\linewidth]{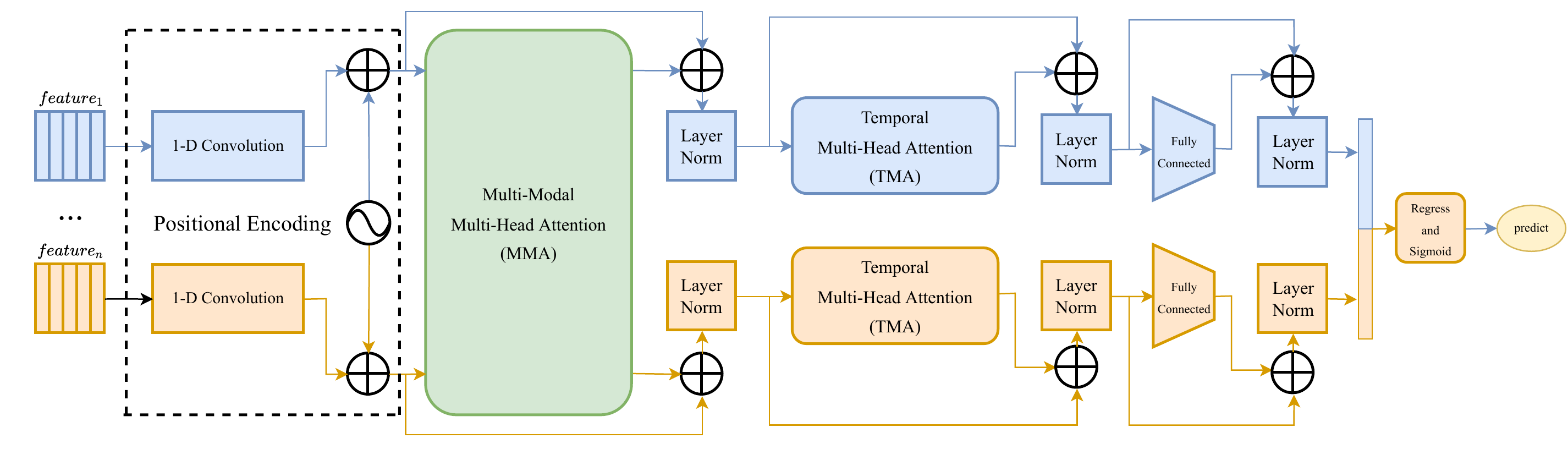}\\
	\caption{The TEMMA framework  \cite{chen2020transformer} used in MuSe-Humor and MuSe-Reaction Sub-Challenges}\label{fig1}
\end{figure*}

In the MuSe-Humor and MuSe-Reaction sub-challenges, the Transformer Encoder with Multimodal Multi-Head Attention (TEMMA) \cite{chen2020transformer} framework is adopted to fusion multimodal features and predict the probability of humor as shown in Figure \ref{fig1}.

\textbf{Feature Embedding}: we use a 1-dimensional temporal convolution network to Capture temporal information for each modal feature vector. For the outputs of the embedded sequence, we add the positional encoding. The length of the final output embedded vector is $d_{model}$.

\textbf{Multimodal Multi-Head Attention (MMA)}: we use the MMA module to obtain complementary information from different modalities and compute inter-modality inter-actions. For a given modality $m \in \left [ 1,M \right ] $ and timestep $t \in \left [ 1,T \right ] $, the input MMA to model inter-modality interactions, which is $Q^t_m=K^t_m=V^t_m$, can be described as Eqn. (\ref{eq1}):
\begin{equation}
\begin{aligned}
& Q^t_m=Concat(Q^t_1 W^Q_1, \dots,  Q^t_m W^Q_m, \dots,  Q^t_M W^Q_M)\\
& K^t_m=Concat(K^t_1 W^K_1, \dots,  K^t_m W^K_m, \dots,  K^t_M W^K_M)\\
& V^t_m=Concat(V^t_1 W^V_1, \dots,  V^t_m W^V_m, \dots,  V^t_M W^V_M)
\label{eq1}
\end{aligned}
\end{equation}
 
\textbf{Temporal Multi-head Attention (TMA)}: TMA is performed individually on each modality to capture the temporal dependency. For a given modality $m \in \left [ 1,M \right ] $ and timestep $t \in \left [ 1,T \right ] $, the input TMA to compute intra-modality dependencies, which is $Q^t_m=K^t_m=V^t_m$, can be described as Eqn. (\ref{eq2}):
\begin{equation}
\begin{aligned}
& Q_m=Concat(Q^1_m , \dots,  Q^t_m, \dots,  Q^T_m) W^Q\\
& K_m=Concat(K^1_m , \dots,  K^t_m, \dots,  K^T_m) W^K\\
& V_m=Concat(V^1_m , \dots,  V^t_m, \dots,  V^T_m) W^V
\label{eq2}
\end{aligned}
\end{equation}

Apart from TMA and MMA, the Multimodal Encoder module contains a residual connection \cite{ba2016layer}, a layer normalization (LN) \cite{mishra2017meta} and a fully connected layer. Finally, we use linear regression and sigmoid function to predict the probability of humor.

\subsection{MuSe-Stress Sub-Challenge}

\begin{figure*}[htb]
	\centering
	\includegraphics[width=0.7\linewidth]{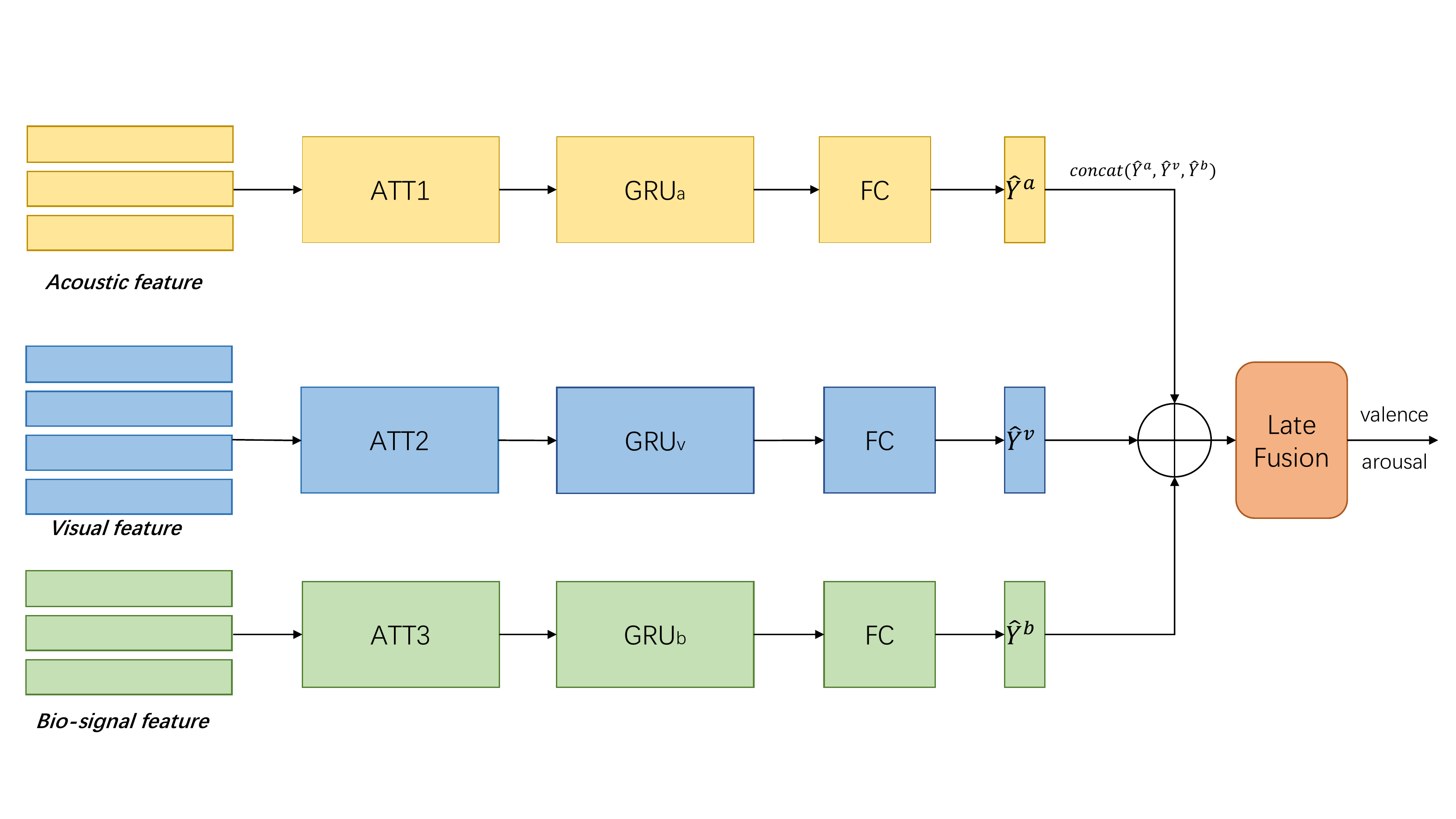}\\
	\caption{The GRU \cite{cho2014properties} with self-attention mechanism used in MuSe-Stress Sub-Challenge}\label{fig2}
\end{figure*}

In the MuSe-Stress sub-challenge, the Gate Recurrent Unit (GRU) \cite{cho2014properties} with the self-attention mechanism is adopted to capture the time-dependent relationship in the time-series features as shown in Figure \ref{fig2}. We take audio, video, and bio-signals features. First, we send the DeepSpectrum feature, ResNet-18 feature and bio-signals feature into the model respectively. After obtaining the three prediction results, we concatenate the three results in series and send them to the later fusion module for regression.

\textbf{Self-attention Mechanism}: Self-attention is used to capture the relationship between different time series. Input audio sequence $X^a=\{ X^a_i \in \mathbb{R} ^{d_a} | i=1,\dots,\left |T\right| \}  $, output $C^a=\{ C^a_i \in \mathbb{R} ^{d_a} | i=1,\dots,\left |T\right| \}$ is given by Eqn. (\ref{eq3}):
\begin{equation}
C^a=Softmax(\frac{Q^a K^{aT}}{\sqrt{d_k} } ) V^a 
\label{eq3}
\end{equation}
where $d^a$ is the dimension of the acoustic sequence and $\left |T\right |$ is the max time step. $Q^a$, $V^a$ and $K^a$ represent the queries, values and keys matrix mapped by the input acoustic sequence $X^a$, respectively. $d_k=d_a/h$ is the scale factor and $h$ represents the number of heads. For convenience, the formula can be abbreviated as Eqn. (\ref{eq4}):
\begin{equation}
C^a=ATT_1 (X^a)
\label{eq4}
\end{equation}

Input visual sequence $X^v=\{ X^v_i \in \mathbb{R} ^{d_v} | i=1,\dots,\left |T\right| \}$, output $C^v=\{ C^v_i \in \mathbb{R} ^{d_v} | i=1,\dots,\left |T\right| \}$ is calculated by:
\begin{equation}
C^v=ATT_2 (X^v) 
\label{eq5}
\end{equation}

Input bio-signals sequence $X^b=\{ X^b_i \in \mathbb{R} ^{d_b} | i=1,\dots,\left |T\right| \}$, output $C^b=\{ C^b_i \in \mathbb{R} ^{d_b} | i=1,\dots,\left |T\right| \}$ is calculated by:
\begin{equation}
C^b=ATT_3 (X^b)
\label{eq6}
\end{equation}
where $d^v$ is the dimension of the visual sequence, $d^b$ is the dimension of the bio-signals sequence.

\textbf{GRU}: We use the Gate Recurrent Unit (GRU) to capture the time-dependent relationship in the time-series features. For a given sequence $C$ (can be $C^a$, $C^v$, $C^b$), the gating memory is updated by Eqn. (\ref{eq7}):
\begin{equation}
H_t=GRU(H_{(t-1)},C_t)
\label{eq7}
\end{equation}
where $H_t=\{H_t^i \in \mathbb{R}^h |i=a,v,b\}$, the $t$ indicates the number of times the hidden semantic feature is repeatedly calculated, $GRU=\{GRU_i |i=a,v,b\}$. Then $H^a$, $H^v$ and $H^b$ predict emotion through the full connection layer. For the audio modality, the emotion prediction $\hat{Y} ^a=\{ \hat{Y}_1^a,\dots,\hat{Y}_{\left |T\right |}^a \}$ is given by Eqn. (\ref{eq8}):
\begin{equation}
\hat{Y} ^a=H^a W^a+b^a
\label{eq8}
\end{equation}
where $ W^a \in \mathbb{R} ^{h_{a \times 1}}$, and $b^a$ represents the bias. For the video modality, the emotion prediction $\hat{Y} ^v=\{ \hat{Y}_1^v,\dots,\hat{Y}_{\left |T\right |}^v \}$ is given by:
\begin{equation}
\hat{Y} ^v=H^v W^v+b^v
\label{eq9}
\end{equation}
where $ W^v \in \mathbb{R} ^{h_{v \times 1}}$, and $b^v$ represents the bias. For the bio-signals modality, the emotion prediction $\hat{Y} ^b=\{ \hat{Y}_1^b,\dots,\hat{Y}_{\left |T\right |}^b \}$ is given by:
\begin{equation}
\hat{Y} ^b=H^b W^b+b^b
\label{eq10}
\end{equation}

\textbf{Late Fusion}: In this sub challenge, we use the late fusion method, concatenate $\hat{Y}^a, \hat{Y}^v, \hat{Y}^b$ as:
\begin{equation}
Y=concat(\hat{Y}^a, \hat{Y}^v, \hat{Y}^b)
\label{eq11}
\end{equation}

\textbf{LSTM}: After late fusion, we use LSTM to capture temporal relationships. For a given sequence $Y$, the gating memory is updated by the following way:
\begin{equation}
H_t=LSTM(H_{t-1},Y_t)  
\label{eq12}
\end{equation}
prediction $\hat{Y}=\{ \hat{Y}_1,\dots,\hat{Y}_{\left |T\right |} \}$ is given by:
\begin{equation}
\hat{Y} =YW+b 
\label{eq13}
\end{equation}
where $W \in \mathbb{R}^h$, and $b$ represents the bias.

\section{Experiments}
\label{exp}


\subsection{Experimental Setup}

In the MuSe-Humor sub-challenge, the TEMMA model structure and hyperparameter settings are basically the same as those used in the MuSe-Reaction sub-challenge, except for the learning rate, which is $1e-3$. We use Binary Cross Entropy loss as a cost function and AUC as an evaluation metric.

In the MuSe-Reaction sub-challenge, we use TMA and TEMMA to conduct the unimodal and multimodal experiments. In the input process block, the number of conv-layers is 5 and the kernel size is 3. The encoder blocks in the Multimodal encoder module is 4 and the number of heads in the multi-head attention layer is 4. For the inference module, the number of nodes in the last fully connected layer is 256 and the dropout is 0.2.

All the experiments are implemented with Pytorch. We adopt the Adam optimizer with the initial learning rate of $1e-4$. All the parameters of optimizer are following the baseline. The learning rate will halve when the training loss does not decrease in 5 continuous epochs, and the training process will early stop when the target can not be optimized for 15 consecutive epochs. We use the Pearson’s Correlations Coefficient ($\rho$) as the evaluation metric.

In the MuSe-Stress sub-challenge, the proposed model consists of the self-attention layer, a bi-directional GRU layer, and the fully connected layer. The number of heads is set to 2, and the number of layers is set to 2 or 4. The number of hidden dimensions of the bi-directional GRU layers is 64 or 128. We set the learning rate 0.001, 0.002, or 0.005, and the AdamW optimizer is used to optimize the whole network. During training, the batch size is set to 256. We train the model for 100 epochs; the learning rate is halved when the loss is not reduced in 15 consecutive epochs. For the late fusion model, we use a Bi-LSTM layer with 6 units to fuse the previous features. The training time for the late fusion model is at most 20 epochs. The learning rate of the AdamW optimizer is 0.002. and the batch size is set to 64.

\subsection{Experimental Results}

For the MuSe-Humor sub-challenge, several unimodal experiments are examined based on multiple features of the three modes, and the experimental validation results are shown in the Table \ref{tab1}. ‘A’, ‘T’ and ‘V’ denote the audio, text and visual modality, respectively.

\begin{table}[!htbp]
\centering
\caption{Unimodal Experimental Results on the validation set of MuSe-Humor Sub-Challenge.}
\label{tab1}
\begin{tabular}{lll}
\hline
Feature                  & Modality & AUC    \\ \hline
eGeMAPS                  & A        & 0.5820 \\
Deep   Spectrum          & A        & \textbf{0.7128} \\
EGeMAPS (Pitch Shift)    & A        & 0.6760 \\
IS09                     & A        & 0.6668 \\
BERT   (sentence level)  & T        & 0.8231 \\
BERT   (word level)      & T        & 0.8069 \\
Phrase   Feature         & T        & \textbf{0.8235} \\
SBERT   (sentence level) & T        & 0.7995 \\
Resnet-18                  & V        & 0.9024 \\
VGGFace2                 & V        & \textbf{0.9223} \\
FAUs                     & V        & 0.9066 \\\hline
\end{tabular}
\end{table}

For acoustic features, the best result is the Deep Spectrum (DS) feature, which achieves an AUC value of 0.7128. For text features, the best result is the phrase feature, which can reach an AUC score of 0.8235. For visual features, the best result is the VGGFace2 feature set, but the other two feature sets are also significantly more effective than the other modal features.

In the multi-modal experiments of the MuSe-Humor Challenge, we try multiple combinations of features and get some results on the validation set. The experimental results are shown in Table \ref{tab2}. We take 4 combinations of feature sets, the first combination (\textbf{Features set 1}) includes eGeMAPS, VGGFace2 and BERT (sentence level). The second combination (\textbf{Features set 2}) is Deep Spectrum, eGeMAPS, eGeMAPS (Pitch Shift), BERT (sentence level), BERT (word level), VGGFace2 and ResNet-18 feature. The third set of feature combinations (\textbf{Features set 3}) adds additional FAUs and IS09 features on top of the second set of features, and the fourth combination (\textbf{Features set 4}) contains Deep Spectrum, phrase feature, BERT (sentence level), Resnet-18, VGGFace2 and FAUs. The third combination of feature sets achieves the best results on the validation set. However, we also find the problem of overfitting in our experiments, which may be due to the model learning the individual features of the samples in the data.

\begin{table}[!htbp]
\centering
\caption{Multi-modal Experimental Results on the validation set of MuSe-Humor Sub-Challenge.}
\label{tab2}
\begin{tabular}{lll}
\hline
Features         & AUC    \\ \hline
Features set 1  & 0.9343 \\
Features set 2 & 0.9467 \\
Features set 3   & \textbf{0.9546} \\
Features set 4 & 0.9514 \\\hline
\end{tabular}
\end{table}

For the MuSe-Reaction sub-challenge, we conduct several unimodal experiments based on the visual modality and audio modality. The results are shown in Table \ref{tab3}. The result for the best of 5 fixed seeds is given.

\begin{table}[!htbp]
\centering
\caption{Unimodal results on the validation set of MuSe-Reaction Sub-Challenge.}
\label{tab3}
\begin{tabular}{lll}
\hline
Feature      & Modality & P       \\ \hline
IS09         & A        & 0.0682  \\
IS13         & A        & 0.1037  \\
eGeMAPS      & A        & 0.0733  \\
MFCCs        & A        & 0.1043  \\
CNN14        & A        & 0.15817 \\
DeepSpectrum & A        & \textbf{0.1835}  \\
VGGFace2     & V        & 0.2727  \\
FAUs         & V        & 0.3107  \\
Resnet-18       & V        & \textbf{0.3893}  \\\hline
\end{tabular}
\end{table}

For the audio modality, the feature that performs best on the validation set is DeepSpectrum, which is relatively only matched by CNN14. Due to the limited information and various noise in the audio modality, this may lead to poor results for low-level features such as eGeMAPS, IS09, IS13, and MFCCs. While DeepSpectrum is based on speech spectrograms and CNN14 are the high-level deep acoustic representations that capture more general acoustic scenes and non-speech descriptions and they are much better. For the visual modality, FAUs with TMA model get better performance than the baseline model. Since VGGFace2 is extracted by the face recognition model, it does not have a close relationship with facial expression, so the effect is general. As for \textbf{ResNet-18} \textsl{which has excellent recognition accuracy pretrained on AffectNet}, it can capture the semantic information of expressions more comprehensively, which makes the visual feature extracted by it could get superior performance than VGGFace2.

In the multimodal experiments of MuSe-Reaction, we combine the best performing feature in unimodal experiments. The results are shown in Table \ref{tab4}.

\begin{table}[!htbp]
\centering
\caption{Multimodal results on the validation set of MuSe-Reaction Sub-Challenge.}
\label{tab4}
\begin{tabular}{ll}
\hline
Features                       & P      \\ \hline
Resnet-18 + eGeMAPS                 & 0.3809 \\
Resnet-18 + CNN14                   & 0.3839 \\
Resnet-18 + DeepSpectrum            & \textbf{0.3968} \\
Resnet-18 + DeepSpectrum + FAUs       & 0.3929 \\
Resnet-18 + DeepSpectrum + FAUs + CNN14 & 0.3930 \\\hline
\end{tabular}
\end{table}

The ResNet-18 feature combined with low-level audio feature such as eGeMAPS and CNN14 make the results worse than the performance of Resnet-18 feature, which is the same as the baseline’s multimodal fusion experimental results. The combination of the best visual and audio performing features, Resnet-18 for facial expression recognition (Resnet) and DeepSpectrum (DS) can achieve better results. This suggests that the two features can complement each other to provide more comprehensive expressive information. It does not yield beneficial results when we add the FAUs or other features to the combination of Resnet-18 and DeepSpectrum.

For the MuSe-Stress sub-challenge, we evaluate the performance of each modality we used. To verify the effectiveness of the proposed model, we conduct the following experiments. The experiment results are given in Table \ref{tab5}.  ‘A’, ‘V’, ‘T’ and ‘B’ represent the audio, video, text, and bio-signal modality. 

\begin{table}[!htbp]
\centering
\caption{ Unimodal results in CCC  on the development set of MuSe-Stress Sub-Challenge.}
\label{tab5}
\begin{tabular}{llll}
\hline
Feature     & Modality & Arousal & Valence \\ \hline
eGeMAPS    & A        & 0.3739  & 0.5144  \\
DS          & A        & 0.4424  & 0.5882  \\
FAUs        & V        & 0.5250  & 0.4587  \\
Resnet-18        & V        & 0.4051  & 0.5436  \\
vggface2    & V        & 0.2981  & 0.2089  \\
BERT-4      & T        & 0.3126  & 0.3698  \\
Bio-signals & B        & 0.3525  & 0.4380  \\\hline
\end{tabular}
\end{table}

In audio modality, the DS feature makes better performance. In video modality, the ResNet-18 feature gets the best performance on valence, feature FAUS gets the best performance on arousal. In addition, audio, video and bio-signal feature performs better than text.

In our approach, we use the late fusion strategy. Table \ref{tab6} shows the CCC performance of different modalities on the developed set of MuSe-Stress sub-challenge using the late fusion strategy.

\begin{table} [!htbp]  \small
\centering
\caption{Different modalities using late fusion strategy on the development set of MuSe-Stress Sub-Challenge (CCC performance).}
\label{tab6}
\begin{tabular}{p{3.9cm}|lll}
\hline
Feature                        & Modality & Arousal & Valence \\ \hline
DS + FAUs/ResNet-18                    & A+V      & 0.3433  & 0.6492  \\
DS + FAUs/ResNet-18 + Bio-signals       & A+V+B    & 0.6364  & 0.7755  \\
DS + FAUs/ResNet-18 + BERT-4 + Bio-signals & A+V+T+B  & 0.4297  & 0.6297   \\\hline
\end{tabular}
\end{table}

From this table, we can find that the multimodal features work better than unimodal features, and the bio-signal features are best fused with audio and video features, which can improve the performance of models based on audio and video feature fusion. The best results for predicting arousal and value are 0.6364 and 0.7755, respectively. Table \ref{tab7} shows the best submission results of our method in stress sub-challenges. Our proposed method is 0.0566 higher than the baseline in combined performance.

\begin{table}[!htbp]
\centering
\caption{The best submission results of our proposed method on the test set of MuSe-Stress Sub-Challenge.}
\label{tab7}
\begin{tabular}{llll}
\hline
Model    & Arousal & Valence & Combined \\ \hline
Baseline & 0.4761  & 0.4931  & 0.4585   \\
Our      & 0.5549  & 0.5857  & 0.5151   \\\hline
\end{tabular}
\end{table}

\section{Conclusion}
\label{con}

In this paper, we present our solutions for the Multimodal Sentiment analysis challenge (MuSe) 2022. For the MuSe-Humor sub-challenge, we used Phrase feature, BERT (sentence level) feature, ResNet-18 feature, VGGFace2 feature and FAUs for model training, and we chose the TEMMA model to accomplish the fusion and judgment of multimodal features. Our model can reach the AUC score of 0.8932 on the test set. For the MuSe-Reaction sub-challenge, we use audio features involving IS09, IS13, eGeMAPS, CNN14 and DeepSpectrum and visual features involving FAUs and Resnet-18. Besides, we adopt the TMA and TEMMA models to conduct unimodal and multimodal experiments. Our approach significantly outperforms the baseline, and the Pearson's correlations coefficient on the test set is 0.3879, outperforming other participants. For the MuSe-Stress sub-challenge, our model takes the features of DS, ResNet-18, FAUs and Bio-signals as inputs, after bidirectional GRUs with attention, and we use the late fusion strategy. The results show our approach outperforms the baseline of 0.0788 in arousal and 0.0926 in valence on the test dataset, obtaining the final combined result of 0.5151.

\bibliographystyle{ACM-Reference-Format}
\balance
\bibliography{sample-base}

\end{document}